\documentclass[runningheads]{llncs}

\usepackage[year=2026]{accv}
\usepackage{accvabbrv}

\usepackage{graphicx}
\usepackage{booktabs}
\usepackage{amsmath,amssymb}
\usepackage{subcaption}
\usepackage{multirow}
\usepackage{tikz}

\usepackage{iftex}
\ifpdftex
  \usepackage[accsupp]{axessibility}  % Improves PDF readability for those with disabilities.
\fi

\usepackage[pagebackref,breaklinks,colorlinks,citecolor=accvblue]{hyperref}

\usepackage{orcidlink}

\begin{document}

% ---------------------------------------------------------------
\title{Unveil the Mask:\\Jointly-Optimized Diffusion Masked Pretraining for Appearance-Agnostic Operator Verification} 

\titlerunning{Unveil the Mask: Jointly-Optimized Diffusion Masked Pretraining}

\author{Katsuya Ogata\inst{1} \and
Ryoya Yoshiyama\inst{1} \and
Kawachi Hodaka\inst{1} \and
Yuta Nakashima\inst{1}}

\authorrunning{K.~Ogata, R.~Yoshiyama, H.~Kawachi, and Y.~Nakashima}

\institute{The University of Osaka\\
\email{\{ogata, yoshiyama, kawachi, n-yuta\}@im.sanken.osaka-u.ac.jp}}

\maketitle

\begin{abstract}
Photorealistic avatar videos decouple the appearance of a rendered face from the person who operates it, which makes conventional face recognition unsuitable for identifying the operator.
Verifying the operator therefore requires representations of how a face moves that are insensitive to what the avatar looks like.
We propose a verification framework built without any external pretrained model.
It uses two complementary modalities provided for each sample: facial landmark sequences and video frames.
From the landmark modality, it learns motion representations by diffusion masked pretraining over grouped landmarks, guided by lightweight auxiliary tasks that predict behavioral cues such as motion intensity and eye openness, and turns them into operator embeddings with a contrastive linear probe.
To generalize to unseen avatar generators, the pretraining input is randomly transformed to simulate how a generator may alter the operator's motion.
From the video modality, it captures fine facial motion through differences between consecutive frames.
The embeddings of the two modalities are combined into one.
Our approach reaches an overall AUC of 0.769 on the WBA2026 benchmark.
Linear probing shows that the video embedding retains little information about the avatar's identity, whereas the landmark embedding retains some, and fusing the two modalities improves verification in particular across speaking and listening modes.
\keywords{Operator Verification \and Diffusion Masked Pretraining}
\end{abstract}

% ===============================================================
\section{Introduction}
\label{sec:intro}

The rapid evolution of face reenactment (animating one person's face with another person's facial motion) has enabled photorealistic 2D and 3D avatars in telepresence, virtual reality, online education, and AI-mediated communication.
However, decoupling facial appearance from the physical operator opens unprecedented security vulnerabilities.
Malicious actors can manipulate avatars to impersonate legitimate operators, rendering traditional biometric authentication, such as image-based face recognition, fundamentally obsolete, as the observed texture and facial geometry belong to the synthesized avatar rather than the operator behind it.

The Who is Behind the Avatar (WBA) task introduces the challenge of appearance-agnostic operator identity verification: identifying the operator, called the \emph{driver} in the task, regardless of the avatar being animated, called the \emph{source}.
This task presents three critical difficulties:
\begin{enumerate}
    \item \textbf{Appearance as a Strong Confounder:} The static geometry and visual texture of the source identity constitute dominant signals in standard neural encoders, making models prone to shortcut learning where identity verification collapses into avatar matching.
    \item \textbf{Distinct Speaking and Listening Behaviors:} Drivers exhibit different movement characteristics between active speech (\emph{speaking} mode, dominated by large-amplitude lip and jaw kinematics) and passive reception (\emph{listening} mode, dominated by subtle micro-expressions, gaze shifts, and head nodding).
    \item \textbf{Cross-Generator Generalization:} Maintaining invariant driver signatures is highly challenging when avatars are generated across distinct rendering engines and unseen generation pipelines.
\end{enumerate}

To address these challenges, we present a self-contained framework that learns appearance-invariant dynamic behavioral signatures without external pretrained weights or distillation.
Our framework combines diffusion masked pretraining on appearance-normalized landmark tokens and lightweight auxiliary behavioral tasks with a driver-aware contrastive linear probe on the frozen encoder, in which every sample has negatives rendered on the same avatar.
To prepare the landmark branch for unseen generators, we randomly transform the landmark motion during pretraining, simulating how a generator may alter it.
We fuse this landmark branch with a video branch that learns from differences between consecutive frames of landmark-guided face crops.
Our final system reaches an overall AUC of 0.769 on the WBA2026 test sets.

% ===============================================================
\section{Related Work}
\label{sec:related}

\subsection{Behavioral Biometrics and Facial Dynamics}
Behavioral biometrics identify individuals by characteristic motion patterns rather than static appearance.
Gait recognition is a representative example, with extensive work from Takemura et al. demonstrating person verification under nuisance factors such as viewpoint changes and unconstrained video conditions~\cite{takemura2019crossview,makihara2021gait}.
More recently, continuous face authentication has extended biometric recognition to streaming facial observations, explicitly suppressing identity-irrelevant variations over time~\cite{li2025continuous,li2026ouca}.
Beyond static facial appearance, prior studies have shown that head and mouth motion, facial-expression dynamics, action-unit statistics, and even micro-expressions contain person-specific behavioral cues~\cite{saeed2006head,benedikt2008facial,kashyap2012facial,saeed2021micro}.
Motivated by these findings, we represent identity through dense landmark trajectories, focusing on dynamic facial behavior while minimizing dependence on avatar appearance.

\subsection{Self-Supervised Predictive and Masked Modeling}
Prediction-based pretraining has become a general mechanism for learning transferable representations without task-specific labels.
Autoregressive language models learn from next-token prediction~\cite{radford2018improving}, while BERT introduced masked-token prediction to exploit bidirectional context~\cite{devlin2019bert}.
The same principle was subsequently extended to visual data through masked autoencoding~\cite{he2022mae}, and to temporal and articulated signals through VideoMAE~\cite{tong2022videomae}, MotionBERT~\cite{zhu2023motionbert}, and masked motion prediction~\cite{mao2023mamp}.
These objectives encourage representations governed by the intrinsic structure of the input rather than a particular downstream decision boundary.
Most recently, Diffusion Masked Pretraining (DiMP)~\cite{dimp} combines masked modeling with diffusion to robustly infer missing spatio-temporal structure.
We adapt this principle to facial landmark sequences.

% \subsection{Joint Optimization of Auxiliary Objectives}
% Multi-task learning has long been studied as a form of inductive transfer in which related objectives regularize a shared representation~\cite{caruana1997multitask}.
% This idea also extends to self-supervised learning, where combining multiple pretext tasks can improve representation transfer~\cite{doersch2017multitask}.
% More recently, VGGT~\cite{wang2025vggt} demonstrates that jointly supervising partially overlapping geometric quantities including camera parameters, depth, and tracking improves the shared representation and downstream point-map estimation.
% However, joint objectives may differ in scale or produce conflicting gradients, motivating adaptive loss weighting~\cite{kendall2018multitask} and gradient surgery~\cite{yu2020pcgrad}; similar interference has been observed when combining masked prediction and contrastive objectives for skeleton motion~\cite{zhang2023pcm3}.
% Accordingly, we use low-weight, label-free behavioral objectives to complement diffusion pretraining while reserving identity supervision for subsequent fine-tuning.

% ===============================================================
\section{Task Formulation}
\label{sec:formulation}

The WBA task~\cite{wba2026_website} asks whether two avatar videos are driven by the same person, while the rendered avatars may differ.

\subsection{Problem Setting}
\label{sec:setting}
Each sample $\vec{x} = (\vec{L}, \vec{V})$ is an avatar video with two synchronized observations:
\begin{itemize}
    \item a 2D facial landmark sequence $\vec{L} \in \mathbb{R}^{T \times K \times 2}$ of $T$ frames and $K$ landmarks ($K = 106$), given in the pixel coordinates of the rendered frames; and
    \item the rendered video $\vec{V} \in \mathbb{R}^{T \times H \times W \times 3}$ with $H = W = 512$.
\end{itemize}
Every driver-source pair contributes two speaking and two listening samples.
Test samples carry none of these labels; in particular, the mode of a test sample is unknown at inference time.
Furthermore, the test set is divided into two subsets to evaluate cross-generator generalization: \textbf{Same Gen} consists of avatars generated by the same rendering pipelines as the training set, while \textbf{Diff Gen} features avatars rendered by distinct, unseen generation pipelines.

The number $T$ of frames is at most 100, and almost all samples have $T = 100$ (98.1\% of the training samples, 98.5\% of Same Gen and all of Diff Gen). The shortest samples have $T = 33$ (training) and $T = 5$ (Same Gen).

\Cref{tab:splits} summarizes the dataset.
For model selection, we hold out 20 drivers and 12 sources that do not appear in our training split, and use the samples of these identities as the validation set.

\begin{table}[t]
\centering
\caption{\textbf{Dataset Statistics.} The validation set is held out from the official training data; its drivers and sources do not occur in our training split.}
\label{tab:splits}
\begin{tabular}{lrrr}
\toprule
Split & Samples & Drivers & Sources \\
\midrule
Official training data & 89,133 & 201 & 111 \\
\quad Our training split & 71,577 & 181 & 99 \\
\quad Our validation split & 960 & 20 & 12 \\
Test, Same Gen (unlabeled) & 40,200 & -- & -- \\
Test, Diff Gen (unlabeled) & 40,200 & -- & -- \\
\bottomrule
\end{tabular}
\end{table}

The objective of this task is to build an encoder $f$ that maps each sample to a unit-norm embedding
\begin{equation}
    \vec{z} = f(\vec{x}) \in \mathbb{R}^{D}, \qquad \|\vec{z}\|_2 = 1, \label{eq:sim}
\end{equation}
such that the cosine similarity $\cos(\vec{z}_i, \vec{z}_j) = \vec{z}_i \vec{z}_j^\top$ is high if $d_i = d_j$ and low otherwise, regardless of $s_i$ and $s_j$, for samples $i$ and $j$.
The challenge protocol accepts only one embedding per test sample, together with the choice of a cosine similarity or Euclidean distance (\cref{eq:sim} is for cosine similarity).

\subsection{Evaluation Metrics}
\label{sec:metrics}
Test samples are divided into probes and galleries, and verification is scored by the area under the ROC curve (AUC) over probe-gallery pairs.
AUCs are computed separately along two axes:
\begin{itemize}
    \item \textbf{Mode combination} of probe and gallery: S$\to$S, S$\to$L, L$\to$S and L$\to$L (S: speaking, L: listening).
    \item \textbf{Generator condition}: pairs rendered by the same generator (\emph{Same Gen}) and by different generators (\emph{Diff Gen}).
\end{itemize}
The Same Gen and Diff Gen scores are each the mean of the four mode-combination AUCs, and the overall score is the mean of Same Gen and Diff Gen.
On our validation split, which has no generator label, we report a single \emph{mean AUC}: the average of the S$\to$S AUC, the L$\to$L AUC and one cross-mode AUC computed over the S$\to$L and L$\to$S pairs together.

% ===============================================================
\section{Proposed Method}
\label{sec:method}

\begin{figure}[t]
\centering
\includegraphics[width=\linewidth]{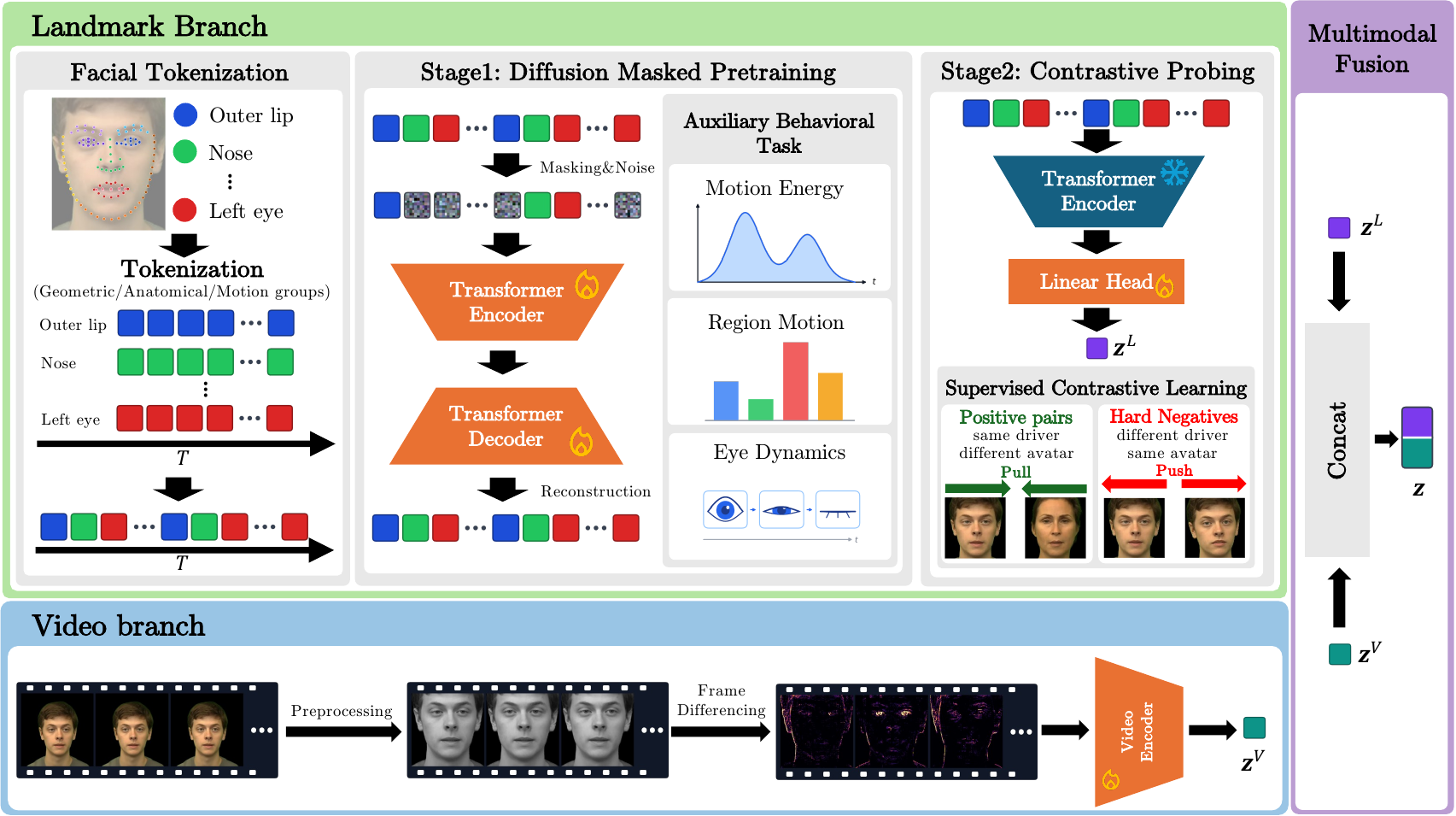}
\caption{\textbf{Overview of our framework.}}
\label{fig:architecture}
\end{figure}

Our framework consists of two branches that are trained independently from scratch (\cref{fig:architecture}).
The \emph{landmark branch} (\cref{sec:landmarks,sec:dimp,sec:multitask,sec:tuning}) learns a motion representation by self-supervised diffusion masked pretraining (Stage 1) and turns it into a driver embedding with a linear contrastive probe on the frozen encoder (Stage 2).
The \emph{video branch} (\cref{sec:video}) learns driver embeddings directly from inter-frame feature differences of landmark-guided face crops.
Several models are ensembled in each branch, and the two branches are fused by weighted concatenation of their normalized embeddings (\cref{sec:fusion}).

\subsection{Appearance-Agnostic Landmark Tokenization}
\label{sec:landmarks}

\paragraph{Removing the avatar's static features.}
The landmark coordinates mix two kinds of information: the static geometry of the rendered face, which belongs to the avatar, and how the face moves over time, which is controlled by the driver.
A model can therefore verify identities by recognizing the avatar's face shape instead of the driver's motion.
We remove the static part with four normalization steps applied to each sample:
\begin{enumerate}
    \item In every frame, the centroid of the 106 landmarks in that frame is subtracted from the coordinates of each landmark, removing the position of the face in the image.
    \item The coordinates in all frames are divided by a single scale, the mean distance of the landmarks to their centroid, removing the size of the face.
    \item For each landmark, its mean position over the frames of the sample is subtracted from its position in every frame, removing the avatar's face shape (\eg, eye spacing and jaw outline). What remains is the displacement of each landmark in each frame from its temporal mean position.
    \item The displacements are divided by their mean Euclidean norm over all frames and landmarks of the sample, so that every sample has the same overall motion amplitude while the relative amplitudes across landmarks and frames are kept.
\end{enumerate}
Sequences shorter than 100 frames are padded by repeating their last frame, and during training we add small Gaussian noise as data augmentation.

\begin{figure}[t]
\centering
\includegraphics[width=\linewidth]{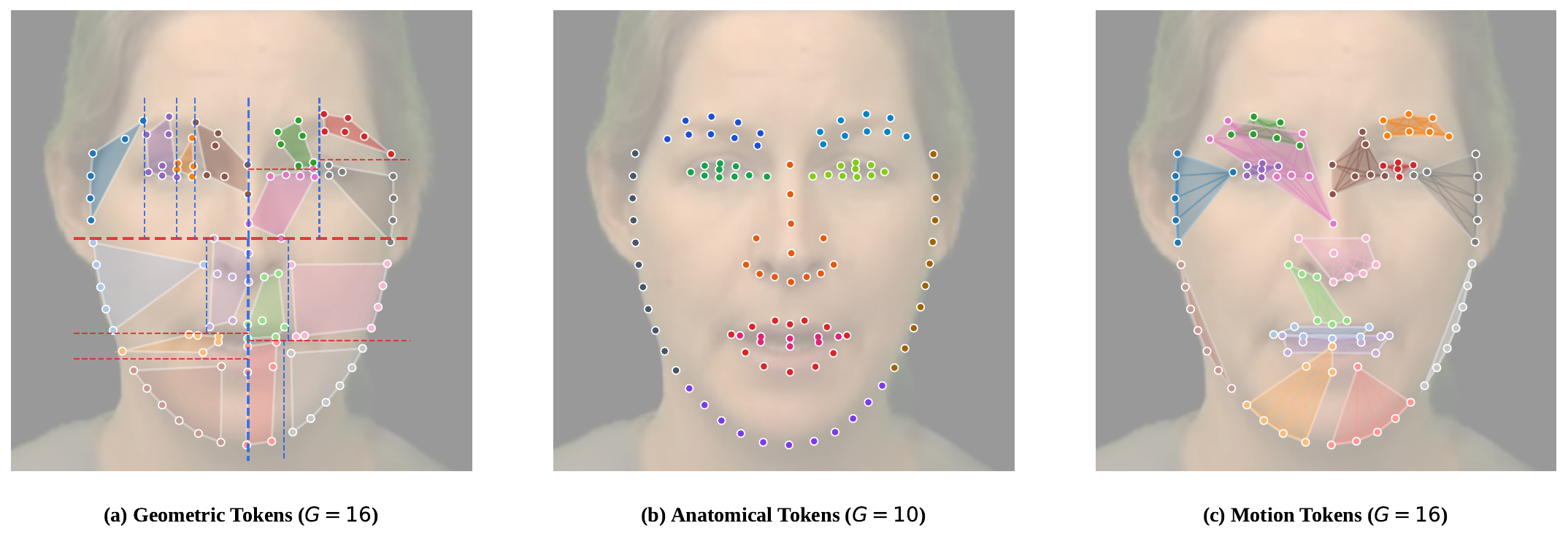}
\caption{\textbf{Comparison of landmark tokenization strategies.}
(a) Geometric tokens ($G = 16$), obtained by recursive median bisection of the mean face shape along its axis of largest spread. Dashed lines illustrate the hierarchical bisection cuts (red: horizontal, blue: vertical), and shaded regions delineate resulting spatially compact clusters that cross anatomical boundaries.
(b) Anatomical tokens ($G = 10$), partitioned by functional facial units on the canonical 106-point topology.
(c) Motion tokens ($G = 16$), obtained via recursive spectral bisection of the landmark displacement correlation graph, grouping landmarks that exhibit synchronized dynamics.}
\label{fig:tokens}
\end{figure}

\paragraph{Spatio-temporal tokens.}
The 106 landmarks are partitioned into $G$ groups, and each group over $l$ consecutive frames ($l = 4$) forms one \emph{spatio-temporal token} (called a \emph{tube} in DiMP~\cite{dimp}), giving $G \times T/l$ tokens per sample.
We use three partitioning strategies (\cref{fig:tokens}):
\begin{itemize}
    \item \textbf{Geometric tokens (Geo):} 16 spatially compact groups ($G = 16$) of 6 to 7 landmarks, obtained by recursive median bisection of the mean face shape along its axis of largest spread (\cref{fig:tokens}a). At each recursion, landmarks are bisected along the coordinate axis of maximum variance. The mean face shape is computed once from the training samples, so the same partition is used for every sample and frame. The resulting clusters (shaded regions in \cref{fig:tokens}a) are purely spatial and ignore facial anatomy, mixing, \eg, eye and jaw landmarks. This yields 400 tokens per sample.
    \item \textbf{Anatomical tokens (Anat):} 10 groups ($G = 10$) defined by a structured partitioning of the 106-point topology into functional facial units (\cref{fig:tokens}b): the jaw contour split into three contiguous segments (11 landmarks each), the left and right eye (10 contour landmarks plus the pupil each), the left and right eyebrow (9 each), the nose (4 bridge and 9 tip landmarks), the outer lip contour (12), and the inner lip contour (8). This preserves natural kinematic boundaries and yields 250 tokens per sample.
    \item \textbf{Motion tokens (Motion):} 16 groups ($G = 16$) of landmarks that move together (\cref{fig:tokens}c). We build a graph over the landmarks whose edge weights are the positive parts of the correlations between the frame-to-frame displacements of two landmarks over the training samples, and split it by recursive spectral bisection into groups of equal size. Like Geo, this yields 400 tokens per sample.
\end{itemize}
% Since the groups contain different numbers of landmarks, every group is padded with dummy landmarks to the size of the largest group; a binary \emph{validity mask} marks the real landmarks, and the padded ones are set to zero in the input and excluded from all losses.
The embedding of a token is the sum of three terms:
(i) a content embedding, a two-layer MLP applied to the flattened coordinates of the token;
(ii) a positional embedding, an MLP applied to the token's centroid and its time index, where the centroid is the mean position of the token's real landmarks over its four frames, measured relative to the mean of all centroids of the sample; it tells the model where on the face and when the token lies; and
(iii) a learnable group embedding, one vector per group, which tells the model which facial part the token belongs to.

\subsection{Diffusion Masked Pretraining (Stage 1)}
\label{sec:dimp}

We adapt Diffusion Masked Pretraining (DiMP)~\cite{dimp} to landmark tokens.
DiMP combines masked modeling with two diffusion processes: one corrupts the masked tokens, whose clean centroids and coordinates must be recovered, and one models their motion.

\paragraph{Masking and corruption.}
Let $\mathcal{T}$ be the set of tokens of a sample, $|\mathcal{T}| = G \times T/l$. Each token $n \in \mathcal{T}$ consists of $K_n$ landmarks observed over $l$ frames. We denote their normalized 2D positions by $\vec{P}_n \in \mathbb{R}^{l \times K_n \times 2}$, their centroid by $\vec{c}_n \in \mathbb{R}^2$, their displacements between consecutive frames by $\vec{D}_n \in \mathbb{R}^{(l-1) \times K_n \times 2}$, and by $\phi^\text{enc}_n(\vec{c})$ the sum of the positional embedding of a centroid $\vec{c}$ at the token's time index and the token's group embedding (\cref{sec:landmarks}).
For each sample, a random subset $\mathcal{M} \subset \mathcal{T}$ of 60\% of the tokens is masked and corrupted, while the other tokens, the \emph{visible} tokens $\mathcal{V} = \mathcal{T} \setminus \mathcal{M}$, remain intact.
As in denoising diffusion models~\cite{ho2020ddpm}, the amount of corruption is set by a timestep $t = 1, \dots, T_\text{d}$, where $T_\text{d} = 1000$,  through a cosine scheduling $\bar{\alpha}_t$, which decreases from nearly 1 (almost no noise) to nearly 0 (almost pure noise) as $t$ grows.
For each sample we draw one such timestep, denoted $t_\text{c}$, following DiMP, uniformly from $1$ to $T_\text{d}$, and share it among all masked tokens; their coordinates and centroids are corrupted with two independent noise samples. That is, for $n \in \mathcal{M}$,
\begin{equation}
    \vec{P}_n^{t_\text{c}} = \sqrt{\bar{\alpha}_{t_\text{c}}}\, \vec{P}_n + \sqrt{1 - \bar{\alpha}_{t_\text{c}}}\, \boldsymbol{\epsilon}_n, \qquad
    \vec{c}_n^{t_\text{c}} = \sqrt{\bar{\alpha}_{t_\text{c}}}\, \vec{c}_n + \sqrt{1 - \bar{\alpha}_{t_\text{c}}}\, \boldsymbol{\epsilon}'_n,
    \label{eq:noise}
\end{equation}
where $\boldsymbol{\epsilon}_n \in \mathbb{R}^{l \times K_n \times 2}$ and $\boldsymbol{\epsilon}'_n \in \mathbb{R}^{2}$ have the same shapes as $\vec{P}_n$ and $\vec{c}_n$, and each of their elements is drawn independently from the standard Gaussian distribution.
\paragraph{Encoder.}
The encoder, a Transformer with hidden dimension $h$, keeps the visible and masked tokens in separate streams.
Each visible token is embedded from its clean coordinates and each masked token from its corrupted ones,
\begin{equation}
    \vec{e}_n =
    \begin{cases}
        \phi^\text{v}(\vec{P}_n) + \phi^\text{enc}_n(\vec{c}_n), & n \in \mathcal{V}, \\
        \phi^\text{m}(\vec{P}_n^{t_\text{c}}) + \phi^\text{enc}_n(\vec{c}_n^{t_\text{c}}), & n \in \mathcal{M},
    \end{cases}
    \label{eq:token-embed}
\end{equation}
where $\phi^\text{v}$ and $\phi^\text{m}$ are separate embedding networks for clean and corrupted coordinates; masked tokens are thus placed at their corrupted centroids.
Stacking the embeddings $\vec{e}_n$ as rows gives the inputs of the two streams at layer $\ell = 0$, $\vec{Z}^\text{v}_{\ell} \in \mathbb{R}^{|\mathcal{V}| \times h}$ for the visible tokens and $\vec{Z}^\text{m}_{\ell} \in \mathbb{R}^{|\mathcal{M}| \times h}$ for the masked tokens.
In layer $\ell$, with $\mathrm{Attn}(\boldsymbol{Q}, \boldsymbol{K}, \boldsymbol{V}) = \mathrm{softmax}(\boldsymbol{Q}\boldsymbol{K}^\top / \sqrt{h})\,\boldsymbol{V}$, the two streams are updated as
\begin{align}
    \vec{Z}^\text{v}_{\ell+1} &= \vec{Z}^\text{v}_{\ell} + \mathrm{Attn}\bigl(\vec{Z}^\text{v}_{\ell} \vec{W}_\text{Q},\ \vec{Z}^\text{v}_{\ell} \vec{W}_\text{K},\ \vec{Z}^\text{v}_{\ell} \vec{W}_\text{V}\bigr), \label{eq:attn-vis} \\
    \vec{Z}^\text{m}_{\ell+1} &= \vec{Z}^\text{m}_{\ell} + \mathrm{Attn}\bigl(\vec{Z}^\text{m}_{\ell} \vec{W}'_\text{Q},\ \vec{Z}^\text{v}_{\ell} \vec{W}'_\text{K},\ \vec{Z}^\text{v}_{\ell} \vec{W}'_\text{V}\bigr), \label{eq:attn-mask}
\end{align}
where the feed-forward network and layer normalization of each layer are omitted for brevity.
The outputs of the last layer, after a final layer normalization, are denoted $\vec{Z}^\text{v}$ and $\vec{Z}^\text{m}$, and their rows by $\vec{z}^\text{v}_n$ and $\vec{z}^\text{m}_n$.
Masked tokens only issue queries: their keys and values come from the visible tokens, so a corrupted token is inferred from the clean context alone and never from other corrupted tokens.
Conversely, visible tokens never attend to masked ones, so the visible stream is unaffected by the corruption.

\paragraph{Centroid prediction.}
A small MLP predicts the clean centroid $\hat{\vec{c}}_n$ of each masked token, \ie, the mean position of its $K_n$ landmarks over its $l$ frames, from the encoder output $\vec{z}^\text{m}_n$, with the loss
\begin{equation}
    \mathcal{L}_{\text{cen}} = \frac{1}{|\mathcal{M}|} \sum_{n \in \mathcal{M}} \bigl\lVert \hat{\vec{c}}_n - \vec{c}_n \bigr\rVert_2^2 .
    \label{eq:cen}
\end{equation}

\paragraph{Decoder and reconstruction.}
The decoder $\psi$, a stack of self-attention layers, takes the visible and masked tokens as one sequence,
\begin{align}
    \vec{d}^\text{v}_n &= \vec{z}^\text{v}_n + \phi^\text{dec}_n(\vec{c}_n) + \boldsymbol{\tau}(t_\text{c}), && n \in \mathcal{V}, \label{eq:dec-vis} \\
    \vec{d}^\text{m}_n &= \phi^\text{m}(\vec{P}_n^{t_\text{c}}) + \phi^\text{dec}_n\bigl(\mathrm{sg}(\hat{\vec{c}}_n)\bigr) + \boldsymbol{\tau}(t_\text{c}), && n \in \mathcal{M}, \label{eq:dec-mask} \\
    \{\vec{u}_n\}_{n \in \mathcal{T}} &= \psi\bigl(\{\vec{d}^\text{v}_n\}_{n \in \mathcal{V}} \cup \{\vec{d}^\text{m}_n\}_{n \in \mathcal{M}}\bigr), \label{eq:decoder}
\end{align}
where $\phi^\text{dec}_n$ is the positional and group embedding of the decoder, of the same form as $\phi^\text{enc}_n$ but with its own weights; $\mathrm{sg}$ denotes stop-gradient, and $\boldsymbol{\tau}(t_\text{c})$ is an embedding of the timestep $t_\text{c}$.
Visible tokens thus enter as their encoder outputs placed at their true centroids, while masked tokens enter as their corrupted content embeddings placed at the \emph{predicted} centroids, so that the decoder never sees the true positions of the masked tokens.
From the decoder output $\vec{u}_n$ for $n \in \mathcal{M}$, a linear head reconstructs its clean coordinates $\hat{\vec{P}}_n$. Unlike $\mathcal{L}_{\text{cen}}$, the loss $\mathcal{L}_{\text{geo}}$ compares the landmarks frame by frame: with $\vec{P}_n^{(j)} \in \mathbb{R}^{K_n \times 2}$ denoting the $j$-th frame of $\vec{P}_n$,
\begin{equation}
    \mathcal{L}_{\text{geo}} = \frac{1}{|\mathcal{M}|\, l} \sum_{n \in \mathcal{M}} \sum_{j=1}^{l} \mathrm{CD}\bigl(\hat{\vec{P}}_n^{(j)}, \vec{P}_n^{(j)}\bigr),
    \label{eq:geo}
\end{equation}
where $\mathrm{CD}$ is the Chamfer distance between two sets of $K_n$ points.

\paragraph{Motion diffusion.}
A second, independent diffusion process models the displacements $\vec{D}_n$ of the masked tokens.
To cover weak and strong noise evenly, the range $\{1, \dots, T_\text{d}\}$ is split into $\kappa$ consecutive intervals of equal width ($\kappa = 4$), and one timestep is drawn uniformly from each interval, independently of $t_\text{c}$.
At each of these timesteps, noise is added to $\vec{D}_n$ as in \cref{eq:noise}, and a head conditioned on $\vec{u}_n$ predicts the added noise; $\mathcal{L}_{\text{mot}}$ is the resulting noise-prediction error, averaged over the $\kappa$ timesteps and computed over real landmarks.
Optionally, as in classifier-free guidance training~\cite{ho2021cfg}, the condition $\vec{u}_n$ is replaced by a learned null vector for a random subset of the masked tokens during training (\emph{condition dropout}); no guidance is used, since the motion head is discarded after Stage 1.
Because the head is conditioned on $\vec{u}_n$, this loss reaches the encoder through the decoder.

The pretraining loss is
\begin{equation}
    \mathcal{L}_{\text{pre}} = \mathcal{L}_{\text{geo}} + \gamma_{\text{cen}} \mathcal{L}_{\text{cen}} + \lambda_{\text{mot}} \mathcal{L}_{\text{mot}}, \qquad
    \label{eq:pretrain}
\end{equation}
where $\gamma_{\text{cen}}$ and $\lambda_{\text{mot}}$ are loss weights (\cref{sec:experiments}).
As a variant, the reconstruction head can instead predict the displacements $\vec{D}_n$ of the masked tokens; $\mathcal{L}_\text{geo}$ is then replaced by the mean squared error of the displacements, divided by their root mean square in the batch (\emph{velocity reconstruction}).

\subsection{Auxiliary Behavioral Tasks}
\label{sec:multitask}

To steer the encoder towards behavioral cues, three auxiliary regression tasks are added to Stage 1.
Their targets are computed in closed form from the input landmarks, so they require no labels.
Each task has a small MLP head on the encoder's visible-token features, and its gradient reaches the encoder.
\begin{itemize}
    \item \textbf{Motion Energy (ME):} for every visible token, the mean squared inter-frame displacement of its landmarks, predicted from that token's own feature.
    \item \textbf{Region Motion (RM):} for every annotated region except the jaw contour (both eyes, both pupils, both eyebrows, nose bridge, nose tip, outer and inner lip), the mean squared inter-frame displacement over the sample, predicted from the mean of the visible-token features.
    \item \textbf{Eye Shape (ES):} for each eye, the mean and the standard deviation over the sample of its openness, measured as the height-to-width ratio of the bounding box of the eye landmarks (a blink proxy), predicted from the mean of the visible-token features.
\end{itemize}
All three are mean squared errors, added with a small weight:
\begin{equation}
    \mathcal{L}_{\text{Stage1}} = \mathcal{L}_{\text{pre}} + w \left( \mathcal{L}_{\text{ME}} + \mathcal{L}_{\text{RM}} + \mathcal{L}_{\text{ES}} \right), \qquad w = 0.01.
    \label{eq:stage1}
\end{equation}

\subsection{Pseudo-Generator Randomization}
\label{sec:pseudogen}

The generators of Diff Gen are not seen in training, and a different generator may reproduce the driver's motion with different timing, smoothness or tracking noise.
We model a generator as an unknown transformation of the driver's motion and simulate it during Stage 1.
Each training sample is transformed by one of six families, chosen uniformly at random:
(i) identity;
(ii) a smooth global time warp;
(iii) a time shift of each facial part;
(iv) a frequency response of each facial part, either a low-pass filter or a boost of high frequencies;
(v) temporally correlated noise on each landmark, mimicking tracking errors; and
(vi) a composition of two of (ii)--(v).
The facial parts are those of the Anat tokens, and each transformed sample is normalized again by steps 3 and 4 of \cref{sec:landmarks}.
Following GroupDRO~\cite{sagawa2020dro}, the motion loss is computed separately for each family $k$ as $\mathcal{L}_{\text{mot},k}$ and replaced by $\sum_k q_k \mathcal{L}_{\text{mot},k}$, where the weights are updated at every step as $q_k \leftarrow q_k \exp(\eta\, \mathcal{L}_{\text{mot},k})$ and renormalized to sum to one, so that the families whose motion is predicted worst are emphasized.

\subsection{Driver-Aware Contrastive Probing (Stage 2)}
\label{sec:tuning}

In Stage 2 the pretrained encoder is \emph{frozen}, and the masked stream, the decoder and all Stage 1 heads are discarded.
All tokens pass through the visible stream, the token features are averaged, and a single linear layer followed by normalization produces the driver embedding.
Only this linear layer is trained.

The linear layer is trained with the supervised contrastive (InfoNCE) loss~\cite{infonce,supcon} with temperature $\tau$:
\begin{equation}
    \mathcal{L}_{\text{con}} = -\frac{1}{|\mathcal{B}|} \sum_{i \in \mathcal{B}} \frac{1}{|\mathcal{I}(i)|} \sum_{j \in \mathcal{I}(i)} \log \frac{\exp(\vec{z}_i \vec{z}_j^\top / \tau)}{\sum_{\iota \in \mathcal{B} \setminus \{i\}} \exp(\vec{z}_i \vec{z}_\iota^\top / \tau)},
    \label{eq:supcon}
\end{equation}
where $\mathcal{B}$ is a batch, $\vec{z}_i$ is the embedding of sample $i$, and $\mathcal{I}(i) \subset \mathcal{B} \setminus \{i\}$ is the set of the other samples in $\mathcal{B}$ of the same driver as sample $i$.

Batches are built by crossing drivers with avatars.
Each batch contains several drivers and several sources with which all of these drivers have been recorded, and one sample for each (driver, source) combination.
Consequently, every positive pair shows the same driver on \emph{different} avatars, and every sample has negatives rendered on the \emph{same} avatar by different drivers, so matching avatar appearance cannot solve the objective.
For encoders pretrained with pseudo-generator randomization, the same transformations (\cref{sec:pseudogen}) are applied to the Stage 2 input.

\subsection{Video Branch on Landmark-Guided Face Crops}
\label{sec:video}

Landmarks capture only the motion of 106 tracked points, whereas the rendered frames also contain the dense motion, such as that of the cheeks and the skin around the eyes and mouth.
Differencing consecutive frames emphasizes such changes between frames and suppresses the static features of each frame, so that the video branch can provide motion cues complementary to the landmarks.

The video branch follows the inter-frame feature differencing model of Chapariniya~\etal~\cite{chapariniya2026microexpression}, trained from scratch on grayscale frames.

\paragraph{Face crop.}
Shrinking the full $512 \times 512$ frame to the network input leaves the face only about 60 pixels wide, at which point the sub-pixel motion of eyelids and lips is largely lost.
We therefore crop the face using the challenge landmarks (no external face detector).
For each sample, a square box is centered on the bounding box of all landmarks over all frames, with a side proportional to its longer edge.
The same box is used for every frame, so head motion is preserved and the crop itself introduces no motion.
During training only, the box size and center are randomly perturbed per sample.
Regions outside the image are filled with zeros.
The crop is resized to $S \times S$ pixels and normalized per sample to zero mean and unit variance.
We also train models on the full frames without cropping, which keep the head motion relative to the image.

\paragraph{Network and training.}
A per-frame convolutional backbone (a strided convolution stack, a meta-kernel convolution block and a nearest-neighbor edge convolution) maps each frame $\vec{I}_f \in \vec{V}$ to a feature map $F(\vec{I}_f)$.
We use two variants: \emph{feature differencing} differences consecutive feature maps, $F(\vec{I}_{f+1}) - F(\vec{I}_f)$, and \emph{pixel differencing}~\cite{chapariniya2026microexpression} differences the frames before the backbone, $F(\vec{I}_{f+1} - \vec{I}_f)$.
The differences are stacked over time and aggregated by a 3D-convolutional temporal head into an L2-normalized embedding.
The loss is \cref{eq:supcon} plus a weighted cross-source alignment term, the mean cosine distance between embeddings of the same driver on different sources in the batch.
Batches use the same driver--source crossing as Stage 2.

\subsection{Ensembling and Score-Level Fusion}
\label{sec:fusion}

Because Stage 1 depends on the random seed, and models with different tokenizations or input modalities capture different cues, we ensemble $n_\text{L}$ landmark models and $n_\text{V}$ video models.
Given member embeddings $\boldsymbol{z}^{\text{L}}_1, \dots, \boldsymbol{z}^{\text{L}}_{n_\text{L}}$ and $\boldsymbol{z}^{\text{V}}_1, \dots, \boldsymbol{z}^{\text{V}}_{n_\text{V}}$ of a sample, the submitted embedding is their concatenation:
\begin{equation}
    \boldsymbol{z} = \left[ \sqrt{\tfrac{\alpha}{n_\text{L}}}\, \boldsymbol{z}^{\text{L}}_1, \dots, \sqrt{\tfrac{\alpha}{n_\text{L}}}\, \boldsymbol{z}^{\text{L}}_{n_\text{L}},\ \sqrt{\tfrac{1-\alpha}{n_\text{V}}}\, \boldsymbol{z}^{\text{V}}_1, \dots, \sqrt{\tfrac{1-\alpha}{n_\text{V}}}\, \boldsymbol{z}^{\text{V}}_{n_\text{V}} \right],
    \label{eq:fusion}
\end{equation}
which has unit norm, so that the submitted cosine similarity equals
\begin{equation}
    \boldsymbol{z}_i^\top \boldsymbol{z}_j = \frac{\alpha}{n_\text{L}} \sum_{a=1}^{n_\text{L}} {\boldsymbol{z}^{\text{L}}_{a,i}} \boldsymbol{z}^{\text{L}, \top}_{a,j} + \frac{1-\alpha}{n_\text{V}} \sum_{b=1}^{n_\text{V}} {\boldsymbol{z}^{\text{V}}_{b,i} \boldsymbol{z}^{\text{V}, \top}_{b,j}}
\end{equation}
a weighted average of the member similarities.
The fusion needs neither the mode nor any other label of the test sample.

% ===============================================================
\section{Experiments and Results}
\label{sec:experiments}
\newcommand{\tbd}{\textcolor{red}{?}}

\subsection{Experimental Setup}
All models are trained on our training split (\cref{tab:splits}), and all weights are randomly initialized.
The validation split, whose drivers and sources are disjoint from the training split, is used for early stopping of Stage 2 and of the video models, for comparing configurations, and for choosing the fusion weight $\alpha$.

\paragraph{Implementation details.}
In Stage 1, the encoder is a Transformer with 6 layers, hidden dimension $h = 256$ and 8 attention heads, and the decoder has 4 layers of the same width.
All models are optimized with AdamW, gradient clipping at norm 1.0 and a learning rate schedule of linear warmup followed by cosine decay.
Stage 1 is trained for 25,000 iterations with 192 samples per step, a learning rate of $5 \times 10^{-4}$, a weight decay of 0.05 and 1,000 warmup steps.
Stage 2 is trained for up to 5,000 iterations with 96 samples per step, a learning rate of $3 \times 10^{-4}$, a weight decay of 0.05 and 200 warmup steps.
The video models are trained for up to 10,000 iterations with 256 samples per step, a learning rate of $10^{-3}$, a weight decay of $10^{-4}$ and 200 warmup steps.
Stage 2 and the video models are stopped early on the validation mean AUC, and both use a temperature of $\tau = 0.07$.
The loss weights are $\gamma_\text{cen} = 0.1$, $w = 0.01$ and $\lambda_\text{mot} = 1$, or $\lambda_\text{mot} = 10$ for the models with condition dropout (probability 0.15) or pseudo-generator randomization ($\eta = 0.01$); the cross-source alignment term of the video branch has weight 0.1.
Both branches produce 256-dimensional embeddings.
Further details are available in our public code.\footnote{\url{https://github.com/ogatakatsuya/Unveil_The_Mask}}

\subsection{Benchmark Performance on WBA2026}
\label{sec:benchmark_results}

\begin{table}[t]
\centering
\caption{\textbf{Verification performance on the WBA2026 test sets} (official leaderboard, AUC). Same Gen and Diff Gen are the means of their four mode combinations (probe$\to$gallery; S: speaking, L: listening), and Overall is the mean of Same Gen and Diff Gen. Ensembles combine three seeds of each listed configuration. Ours uses different ensembles for Same Gen and Diff Gen (\cref{sec:analysis}).}
\label{tab:benchmark}
\resizebox{\linewidth}{!}{%
\begin{tabular}{l cccc cccc ccc}
\toprule
\multirow{2}{*}{Method} & \multicolumn{4}{c}{Same Gen, mode pairs} & \multicolumn{4}{c}{Diff Gen, mode pairs} & \multirow{2}{*}{Same Gen} & \multirow{2}{*}{Diff Gen} & \multirow{2}{*}{\textbf{Overall}} \\
\cmidrule(lr){2-5} \cmidrule(lr){6-9}
 & S$\to$S & S$\to$L & L$\to$S & L$\to$L & S$\to$S & S$\to$L & L$\to$S & L$\to$L & & & \\
\midrule
Landmark baseline (ST-GCN)~\cite{alsherfawi2025behavioral} & 0.876 & 0.723 & 0.751 & 0.865 & 0.542 & 0.516 & 0.511 & 0.547 & 0.804 & 0.529 & 0.666 \\
Video baseline~\cite{chapariniya2026microexpression} & 0.866 & 0.808 & 0.822 & 0.879 & 0.533 & 0.508 & 0.513 & 0.528 & 0.844 & 0.520 & 0.682 \\
\midrule
\multicolumn{12}{l}{\emph{Landmark branch, single seed}} \\
DiMP (Geo) & 0.813 & 0.635 & 0.653 & 0.727 & 0.639 & 0.561 & 0.563 & 0.644 & 0.707 & 0.602 & 0.654 \\
\quad + auxiliary tasks & 0.841 & 0.664 & 0.686 & 0.767 & 0.653 & 0.568 & 0.566 & 0.656 & 0.739 & 0.611 & 0.675 \\
\midrule
\multicolumn{12}{l}{\emph{Landmark branch, three seeds}} \\
MAE (Geo, no auxiliary tasks) & 0.843 & 0.684 & 0.719 & 0.767 & 0.610 & 0.543 & 0.538 & 0.613 & 0.753 & 0.576 & 0.664 \\
DiMP (Geo, no auxiliary tasks) & 0.826 & 0.652 & 0.678 & 0.755 & 0.644 & 0.569 & 0.554 & 0.654 & 0.728 & 0.605 & 0.667 \\
DiMP (Anat) & 0.875 & 0.708 & 0.739 & 0.790 & 0.655 & 0.567 & 0.570 & 0.654 & 0.778 & 0.611 & 0.695 \\
\quad + pseudo-generator randomization & 0.825 & 0.659 & 0.675 & 0.752 & 0.682 & 0.568 & 0.587 & 0.670 & 0.728 & 0.627 & 0.677 \\
\midrule
\multicolumn{12}{l}{\emph{Video branch, single seed}} \\
Feature differencing~\cite{chapariniya2026microexpression} ($S = 192$) & 0.862 & 0.750 & 0.800 & 0.847 & 0.578 & 0.538 & 0.534 & 0.581 & 0.815 & 0.558 & 0.686 \\
Pixel differencing~\cite{chapariniya2026microexpression} ($S = 128$) & 0.879 & 0.771 & 0.800 & 0.873 & 0.564 & 0.541 & 0.526 & 0.597 & 0.831 & 0.557 & 0.694 \\
\midrule
\multicolumn{12}{l}{\emph{Fusion of the two branches}} \\
Single models (Anat + pixel diff.) & 0.909 & 0.781 & 0.815 & 0.872 & 0.622 & 0.562 & 0.558 & 0.645 & 0.844 & 0.596 & 0.720 \\
\textbf{Ours} & 0.946 & 0.845 & 0.881 & 0.928 & 0.711 & 0.573 & 0.587 & 0.678 & \textbf{0.901} & \textbf{0.637} & \textbf{0.769} \\
\bottomrule
\end{tabular}%
}
\end{table}

\Cref{tab:benchmark} summarizes the results on the test sets.
On landmarks alone, the auxiliary tasks raise the overall AUC of a single DiMP model from 0.654 to 0.675, with gains in all eight mode combinations.
Compared with the landmark baseline, our landmark models are lower on Same Gen but clearly higher on Diff Gen (0.60 to 0.63 vs.\ 0.53, except MAE), \ie, they generalize better to unseen generators.
The video branch alone reaches 0.815 to 0.831 on Same Gen but only about 0.56 on Diff Gen, below the landmark models.
Fusing a single landmark model with a single video model already gives 0.720, above every single-branch result in \cref{tab:benchmark}.
For unseen generators, the diffusion objective and pseudo-generator randomization matter: DiMP scores higher than a masked autoencoder (MAE) with the same tokens and encoder on Diff Gen but lower on Same Gen, and pseudo-generator randomization raises Diff Gen from 0.611 to 0.627 at the cost of Same Gen.
Our final submission, which uses a fused ensemble for Same Gen and a landmark ensemble trained with pseudo-generator randomization for Diff Gen, reaches 0.769 overall.

\subsection{Ablation Studies}
\label{sec:ablations}

\begin{table}[t]
\centering
\caption{\textbf{Ablations on the validation split} (mean AUC, \cref{sec:metrics}). (a) Combinations of the auxiliary tasks motion energy (ME), region motion (RM) and eye shape (ES) with $w = 0.01$ (Geo tokens; ensemble of three seeds); the first row is plain DiMP. (b) Fusion weight $\alpha$ ($\alpha = 1$: landmarks only, $\alpha = 0$: video only) for the ensemble with feature-differencing video models and for the final Same Gen file. (c) Token grouping under an identical recipe (with all three auxiliary tasks; ensembles of three seeds per grouping).}
\label{tab:ablation}
\footnotesize
\begin{subtable}[t]{0.20\linewidth}
\centering
\caption{Auxiliary tasks}
\label{tab:aux-comb}
\begin{tabular}{cccc}
\toprule
ME & RM & ES & AUC \\
\midrule
 & & & 0.811 \\
\checkmark & & & 0.827 \\
 & \checkmark & & 0.820 \\
 & & \checkmark & 0.820 \\
\checkmark & \checkmark & & 0.818 \\
\checkmark & & \checkmark & 0.833 \\
 & \checkmark & \checkmark & 0.808 \\
\checkmark & \checkmark & \checkmark & \textbf{0.834} \\
\bottomrule
\end{tabular}
\end{subtable}%
\hspace{4mm}
\begin{subtable}[t]{0.30\linewidth}
\centering
\caption{Fusion weight $\alpha$}
\label{tab:abl-fusion}
\begin{tabular}{lcc}
\toprule
$\alpha$ & Feat.\ diff. & Final \\
\midrule
1.0 & 0.856 & 0.862 \\
0.9 & 0.891 & 0.906 \\
0.8 & 0.912 & 0.933 \\
0.7 & 0.922 & 0.946 \\
0.6 & \textbf{0.926} & 0.952 \\
0.5 & 0.925 & \textbf{0.952} \\
0.0 & 0.901 & 0.937 \\
\bottomrule
\end{tabular}
\end{subtable}%
\hspace{1mm}
\begin{subtable}[t]{0.20\linewidth}
\centering
\caption{Token grouping}
\label{tab:abl-tokens}
\begin{tabular}{lc}
\toprule
Tokens & AUC \\
\midrule
Geo & 0.834 \\
Anat & 0.850 \\
Motion & 0.854 \\
\midrule
Geo + Anat & 0.856 \\
Geo + Motion & 0.856 \\
Anat + Motion & 0.862 \\
All three & \textbf{0.862} \\
\bottomrule
\end{tabular}
\end{subtable}%
\end{table}

\paragraph{Auxiliary tasks (\cref{tab:aux-comb}).}
Each task alone improves on plain DiMP, and using all three tasks gives the best result (0.834, +2.3 points); ME and ES together are close to it.

\paragraph{Fusion weight (\cref{tab:abl-fusion}).}
We select $\alpha$ on the validation split ($\alpha = 0.6$ and $0.5$, respectively); the fusion is better than either branch alone over a wide range of $\alpha$.

\paragraph{Token grouping (\cref{tab:abl-tokens}).}
Anat and Motion tokens score 1.6 and 2.0 points above Geo tokens.
Combining groupings improves further: every combination is better than any single grouping, and combining all three gives the best result (0.862).

\subsection{Analysis}
\label{sec:analysis}

\paragraph{Appearance invariance via source probing.}
\label{sec:probing}
To measure how much information about the avatar remains in the embeddings, we train a linear classifier on frozen embeddings to predict the source identity and, separately, the mode of a sample.
The classifier is trained on one half of the drivers and evaluated on the other half (15,000 samples of the training split), so that it cannot rely on the driver identity.
We compare single models, whose embeddings all have 256 dimensions.

\begin{table}[t]
\centering
\caption{\textbf{Linear Probing Accuracy of Avatar Source and Mode.} Accuracy of a linear classifier on frozen embeddings, trained and evaluated on disjoint drivers. Chance is the frequency of the most frequent class.}
\label{tab:source_probe}
\resizebox{0.75\linewidth}{!}{%
\begin{tabular}{l @{\hspace{1em}} c @{\hspace{1em}} c @{\hspace{1em}} c}
\toprule
Feature Representation & Source Acc. & Chance & Mode Acc. \\
\midrule
Landmark, Geo tokens & 10.2\% & 1.3\% & 74.7\% \\
Landmark, Anat tokens & 10.3\% & 1.3\% & 75.5\% \\
Video branch & 2.1\% & 1.3\% & 60.3\% \\
\bottomrule
\end{tabular}%
}
\end{table}

As summarized in \cref{tab:source_probe}, the video branch predicts the source only slightly above chance (2.1\%, chance 1.3\%): differencing consecutive frames removes most of the avatar's appearance.
The landmark branch predicts it at about 10\%, well above chance although far from perfect: the normalization removes the avatar's static face shape, but the motion still carries some properties of the rendered avatar.
The landmark embeddings also predict the mode more accurately than the video embeddings.
This analysis uses the training generators only and cannot measure cues that are specific to a generator.

\paragraph{Cross-mode and cross-generator verification.}
Cross-mode pairs are the hardest for the landmark branch: on the validation split, the landmark models of the final Same Gen file reach 0.937 for S$\to$S but 0.817 and 0.850 for S$\to$L and L$\to$S, and their embeddings predict the mode well (\cref{tab:source_probe}), \ie, they retain how the motion of a driver differs between speaking and listening.
The video branch depends less on the mode (0.908 and 0.925 for S$\to$L and L$\to$S), and the fusion raises the cross-mode AUCs to 0.925 and 0.938.
On Diff Gen, however, video models, including the video baseline, score only 0.52 to 0.56 and fall below our landmark models (0.58 to 0.63, \cref{tab:benchmark}), although video is stronger on the validation split, which is rendered by the training generators; the video branch thus appears to rely partly on cues specific to these generators.
We therefore chose to use only landmark models for Diff Gen.

% ===============================================================
\section{Conclusion}
\label{sec:conclusion}
We presented an operator verification framework combining diffusion masked pretraining, auxiliary behavioral tasks, and an inter-frame differencing video branch, achieving 0.769 AUC on WBA2026. While linear probing shows video retains minimal avatar info and fusion helps cross-mode verification, generalizing to unseen generators remains the primary challenge.

\subsubsection*{Acknowledgements.}
This work was partly supported by JSPS KAKENHI Grant Number JP23H00497 and JST FOREST Grant Number JPMJFR216O.

\bibliographystyle{splncs04}
\bibliography{main}
\end{document}